\pdfoutput=1

\documentclass[10pt,twocolumn,letterpaper]{article}

\usepackage{iccv}
\usepackage{times}
\usepackage{epsfig}
\usepackage{graphicx}
\usepackage{amsmath}
\usepackage{amssymb}

\usepackage{savesym}
\savesymbol{checkmark}

\newcommand{\modelName}{InstaNAS }
\newcommand{\modelNamePunc}{InstaNAS} 

\newcommand{\expectation}{\mathop{\mathbb{E}}}

\newcommand{\taskObjective}{\mathop{\mathbb{O}}_{T}}
\newcommand{\archObjective}{\mathop{\mathbb{O}}_{A}}

\usepackage{ulem}
\usepackage{xcolor}
\definecolor{crimson}{rgb}{0.86, 0.08, 0.24}
\definecolor{green}{rgb}{0, 0.5, 0.25}
\definecolor{purple}{rgb}{0.75, 0, 1}
\definecolor{orange}{rgb}{1, 0.5, 0.25}
\definecolor{yellow}{rgb}{1, 1, 0}
\definecolor{new_blue}{rgb}{0, 0.5, 1}
\ifx \submission \undefined

\else

\fi

\usepackage{xspace}
\makeatletter
\DeclareRobustCommand\onedot{\futurelet\@let@token\@onedot}
\def\@onedot{\ifx\@let@token.\else.\null\fi\xspace}

\def\eg{\emph{e.g}\onedot} 
\def\ie{\emph{i.e}\onedot}

\def\etal{\emph{et al}\onedot}
\makeatother

\usepackage{amsmath,amsfonts,bm}

\def\eqref#1{equation~\ref{#1}}

\def\1{\bm{1}}

\DeclareMathAlphabet{\mathsfit}{\encodingdefault}{\sfdefault}{m}{sl}
\SetMathAlphabet{\mathsfit}{bold}{\encodingdefault}{\sfdefault}{bx}{n}

\usepackage[inline]{enumitem}
\usepackage{booktabs}
\usepackage{dingbat}
\usepackage{xcolor}
\usepackage{caption}
\usepackage{authblk}
\usepackage[toc,page]{appendix}
\usepackage{algorithm}
\usepackage{algorithmic}
\usepackage{subfig}
\iccvfinalcopy 

\ificcvfinal\fi
\begin{document}

\title{\modelNamePunc: Instance-aware Neural Architecture Search}

\author[$\dagger$]{An-Chieh Cheng*}
\author[$\dagger$]{Chieh Hubert Lin*}
\author[$\ddagger$]{Da-Cheng Juan}
\author[$\ddagger$]{Wei Wei}
\author[$\dagger$]{Min Sun}

\affil[$\dagger$]{\small National Tsing-Hua University, Hsinchu, Taiwan}
\affil[$\ddagger$]{\small Google Research, Mountain View, USA}



\maketitle

\begin{abstract}
    Conventional Neural Architecture Search (NAS) aims at finding a \textit{single} architecture that achieves the best performance, which usually optimizes task related learning objectives such as accuracy. However, a single architecture may not be representative enough for the whole dataset with high diversity and variety. Intuitively, electing domain-expert architectures that are proficient in domain-specific features can further benefit architecture related objectives such as latency. In this paper, we propose InstaNAS---an instance-aware NAS framework---that employs a controller trained to search for a ``distribution of architectures'' instead of a single architecture; This allows the model to use sophisticated architectures for the difficult samples, which usually comes with large architecture related cost, and shallow architectures for those easy samples. During the inference phase, the controller assigns each of the unseen input samples with a domain expert architecture that can achieve high accuracy with customized inference costs. Experiments within a search space inspired by MobileNetV2 show InstaNAS can achieve up to 48.8\% latency reduction without compromising accuracy on a series of datasets against MobileNetV2.
    \vspace{-1em}
\end{abstract} 

\let\thefootnote\relax\footnotetext{* Equal Contribution}

\section{Introduction}
\label{section:intro}
Neural Architecture Search (NAS) has become an effective and promising approach to automate the design of deep learning models. It aims at finding the optimal model architectures based on their performances on evaluation metrics such as accuracy~\cite{zoph2016neural}. One popular way to implement NAS is to employ reinforcement learning (RL) that trains an RNN controller (or ``agent'') to learn a search policy within a pre-defined search space. In each iteration of the search process, a set of child architectures are sampled from the policy, and evaluate performance on the target task. The performance is then used as the reward to encourage the agent to prioritize child architectures that can achieve a higher expected reward. In the end, a single architecture with a maximum reward will be selected and trained to be the final solution of the task.

\begin{figure}[t]
\begin{center}
\includegraphics[width=1\linewidth]{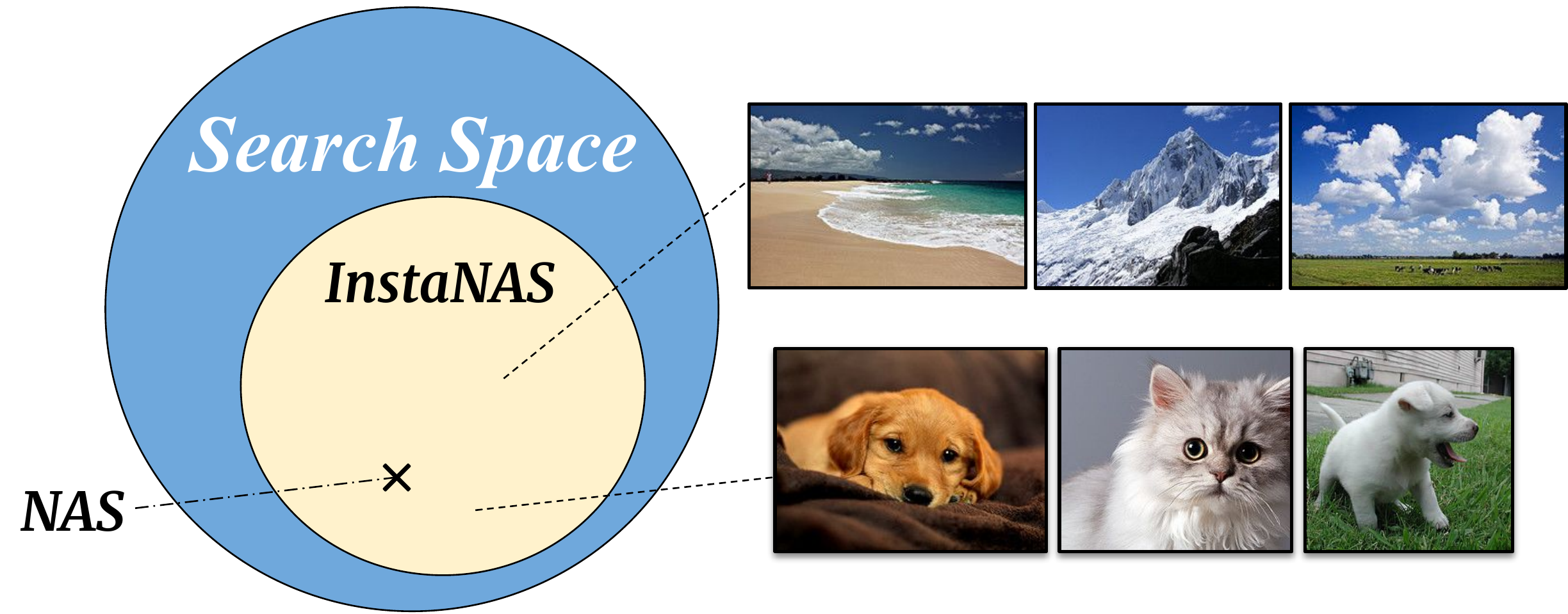} \\ [0.5em]
\includegraphics[width=1\linewidth]{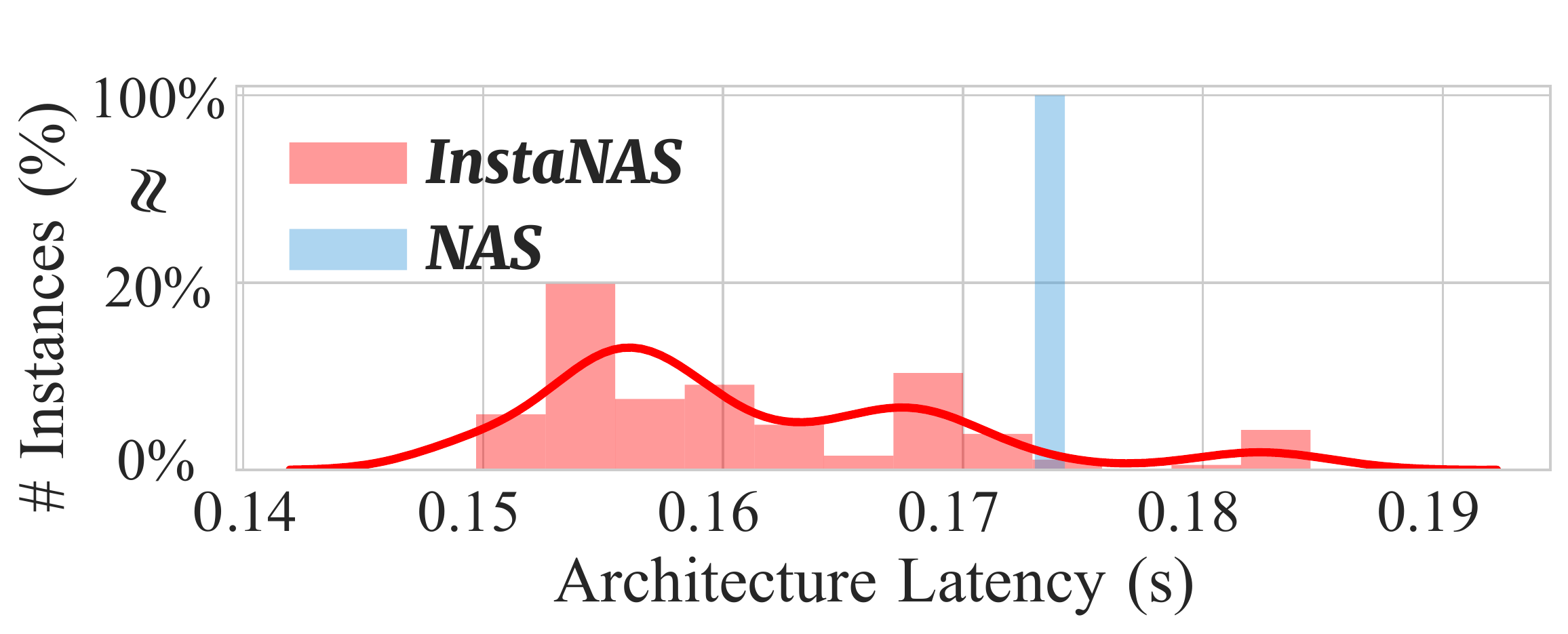}
\end{center}
\caption{InstaNAS searches for a \textbf{\textit{distribution}} of architectures instead of a single one from conventional NAS. We showcase a distribution of architecture latencies found by InstaNAS for CIFAR-10. The InstaNAS controller assigns each input instance to a domain expert architecture, which provides customized latency for different domains of data.}



\label{fig.concept}
\end{figure}




\begin{figure*}[h]
\centering
\includegraphics[width=1\linewidth]{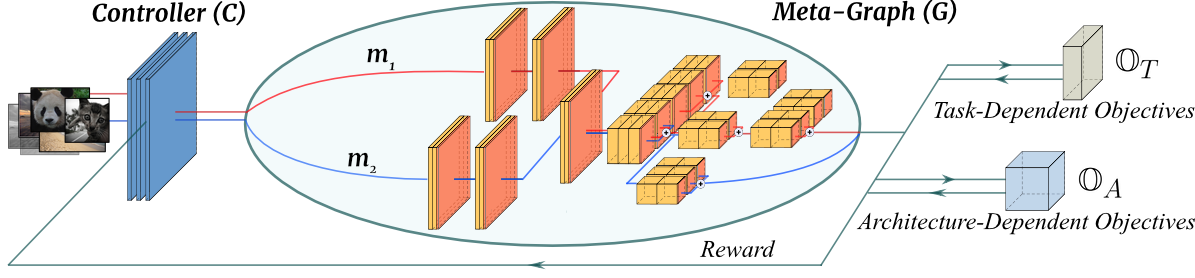}
\caption{InstaNAS controller ($C$) selects an expert child architecture ($m$) from the meta-graph ($G$) for each input instance while considering task-dependent objectives ($\taskObjective$) (e.g., accuracy) and architecture-dependent objectives ($\archObjective$) (e.g., latency).}
\label{fig.flow}

\end{figure*}

Although a single architecture searched using NAS seems to be sufficient to optimize task related metrics such as accuracy, its performance is largely constrained in architecture related metrics such as latency and energy. For example in a multi-objective setting where both accuracy and latency are concerned, NAS is constrained to come up with a single model to explore the trade-off between accuracy and latency for all samples. In practice, however, difficult samples require complicated and usually high latency architectures whereas easy samples work well with shallow and fast architectures.
This inspires us to develop \modelNamePunc, a NAS framework which generates a \textit{\textbf{distribution}} of architectures instead of a single one. Each architecture within the final distribution is an expert of one or multiple specific domains, such as different difficulty, texture, content style and speedy inference. For each sample, the controller is trained to select a suitable architecture from its distribution. With basic components being shared across architectures, weights can be re-used toward architectures that have never been selected before. The \modelName framework allows samples to have their own architectures, making it flexible to optimize architecture related objectives. 



\modelName is critical in many of the recently proposed settings such as multi-objective NAS~\cite{dong2018dpp,hsu2018monas,tan2018mnasnet,yang2018netadapt}, which optimizes not only task-dependent metrics such as accuracy but also those metrics that are architecture-dependent such as latency. In particular, the controller of InstaNAS has the capability of selecting the architectures by considering the variations among instances. 
To enable effective training, we introduce a dynamic reward function to gradually increase the difficulty of the environment, a technique commonly found in curriculum learning. 
In the meanwhile, the reward interval slowly decreases its upper bound through epochs.
Note that InstaNAS also aligns with the concept of conditional computing since the instance-level architecture depends on the given input sample. Most importantly, InstaNAS elegantly combines the ideas of NAS and conditional computing learn a distribution of architectures and a controller to generate instance-level architectures.

In conclusion, the main contributions of this paper are as the following: We propose \modelName, the first instance-aware neural architecture search framework that generates architectures for each individual sample. Instance-awareness allows us to incorporate the variability of samples into account by designing architectures that specifically optimize each sample. To the best of our knowledge, we are the first work toward building NAS with instance-awareness. We show that \modelName is able to out-perform MobileNetV2 dramatically in terms of latency while keeping the same performance. Experimental results illustrate an average of 48.9\%, 40.2\%, 35.2\% and 14.5\% latency reduction with comparable accuracy on CIFAR-10, CIFAR-100, TinyImageNet and ImageNet, respectively. Further latency reduction of 26.5\% can be achieved on ImageNet if a moderate accuracy drop ($\simeq$ 0.7\%) is allowed.

\section{Related Work}
\label{section:relatedwork}

\paragraph{\textbf{Neural Architecture Search.}}

Neural Architecture Search (NAS) has emerged growing interest in the field of AutoML and meta-learning~\cite{vilalta2002perspective} in recent years. Seminal work by Zoph \etal~\cite{zoph2016neural} first proposed “Neural Architecture Search (NAS)” using reinforcement learning algorithm. They introduce a learnable RNN controller that generates a sequence of actions representing a child network within a predefined search space, while the validation performance is taken as the reward to train the controller. Since the process of NAS can also be framed as a natural selection problem, some works~\cite{real2017large,real2018regularized,xie2017genetic} propose to use evolutionary approaches with genetic algorithms to optimize the architecture. However, all these works focus on optimizing model accuracy as their only objective. In real-world, these models may not be suitable for being deployed on certain (\eg, latency-driven) applications, such as mobile applications and autonomous car.

\paragraph{\textbf{Multi-objective Neural Architecture Search.}}
For better flexibility and usability in real-world applications, several works are dedicated to extending NAS into multiple-objective neural architecture search, which attempts to optimize multiple objectives while searching for architectures. Elsken \etal~\cite{elsken2018multi} and Zhou \etal~\cite{zhou2018resource} use FLOPs and the number of parameters as the proxies of computational costs; Kim \etal~\cite{kim17nemo} and Tan \etal~\cite{tan2018mnasnet} directly minimized the actual inference time; Dong \etal~\cite{dong2018dpp} proposed to consider both device-agnostic objectives (\eg, FLOPs) and device-related objectives (\eg, inference latency) using Pareto Optimization. However, all these aforementioned algorithms only consider searching for a single final architecture achieving the best average accuracy for the given task. In contrast, \modelName is an MO-NAS approach that searches for a distribution of architectures aiming to speed up the average inference time with \textit{instance-awareness}. 


\paragraph{\textbf{One-shot Architecture Search.}}
Computational expensive is another fundamental challenge in NAS, conventional NAS algorithms require thousands of different child architectures to be trained from scratch and evaluated, which is often time costly. One-shot architecture search is an approach using share-weight across child architectures to amortize the search cost. The concept of weight sharing has been widely adopted by different NAS approaches with various kinds of search strategies: with evolutionary algorithm~\cite{real2018regularized, real2017large}, reinforcement learning~\cite{pham2018efficient}, gradient descent~\cite{liu2018darts}, and random search~\cite{bender2018understanding}. Instead of training each child architecture from scratch, they allow child architectures to share weights whenever possible. We also adopt the similar design principle of the one-shot architecture search to not only accelerate \modelName but also to reduce the total number of parameters in \modelName. We will explain further detail of how we leverage the one-shot architecture search to build our meta-graph in Section~\ref{section:method:meta-graph}.



\paragraph{\textbf{Conditional Computation.}} 
Several conditional computation methods have been proposed to dynamically execute different modules of a model on a per-example basis~\cite{bengio2015conditional,kuen2018stochastic,liu2017dynamic,teja2018hydranets,wu2018blockdrop,veniat2017learning}. More specifically, Wu \etal~\cite{wu2018blockdrop} use policy network to generate a series of decision which selectively dropped a subset of blocks in a well-known baseline network (\eg, ResNet~\cite{he2016deep}) with respect to each input. However, all methods mentioned above assume their base model to be optimal across all samples, then perform their algorithm as a post-processing method to further reduce computational costs. In contrast, \modelName is a neural architecture search framework with built-in instance awareness during architecture design and inference time decisions. \modelName is relatively a more flexible framework - covering but not limited to all possible solutions which can be provided by previous conditional computation frameworks.


\section{InstaNAS: Instance-aware NAS}
\label{section:method}
In this section, we first give the overview of \modelNamePunc, specifically about the meta-graph and the controller. Then, we describe how the meta-graph is constructed and pre-trained. Finally in Section~\ref{section:method:agent}, we explain how to design the multi-objective reward function for training the controller and updating the meta-graph. We provide a detailed algorithm in the supplementary.

\subsection{Overview}
\label{section:method:overview}

\modelName contains two major components: a meta-graph and a controller. The meta-graph is a directed acyclic graph (DAG), with one source node (where an input image is fed) in the beginning and one sink node (where the prediction is provided) at the end; every node between the source and sink is a computing module such as a convolutional operation, and an edge connects two nodes meaning the output of one node is used as the input of the other. With this meta-graph representation, every path from source to sink can be treated as a valid child architecture as an image classifier. Therefore, the meta-graph can be treated as the set containing all possible child architectures.

The other major component of \modelName is the controller; it is designed and trained to be \textit{instance-aware} and optimize for multi-objective. Particularly, for each input image, the controller selects a child architecture (\ie, a valid path in the meta-graph) that accurately classifies that image, and at the same time, minimizes the inference latency (or other computational costs). Therefore, the controller is trained to achieve two objectives at the same time: maximize the classification accuracy (referred as the task-dependent objective $\taskObjective$) and minimize the inference latency (referred as the architecture-dependent objective $\archObjective$). Note that $\archObjective$ can also be viewed as a constraint when optimizing for the task-dependent objective.

Next, the training phase of \modelName consists of three stages: (a) ``pre-train'' the meta-graph, (b) ``jointly train'' both controller and the meta-graph,  and (c) ``fine-tune'' the meta-graph. In the first stage, the meta-graph (denoted as $G$, parametrized by $\Theta$) is pre-trained with $\taskObjective$. In the second stage, a controller (denoted as $C$, parametrized by $\phi$) is trained to select a child architecture $m(x; \theta_x) = C(x, G; \phi)$ from $G$ for each input instance $x$. In this stage, the controller and the meta-graph are trained in an interleaved fashion: train the controller with the meta-graph fixed in one epoch and vice versa in another epoch. This training procedure enforces the meta-graph to adapt to the distribution change of the controller. Meanwhile, the controller is trained by policy gradient~\cite{sutton2000policy} with a reward function $R$ which is aware of both $\taskObjective$ and $\archObjective$. The training detail of the controller is described in Section~\ref{section:method:agent}. In the third stage, after the controller is trained, we fix the controller and focus on fine-tuning the meta-graph for the task-dependent objective $\taskObjective$; specifically, for each input image the controller selects a child architecture (\ie, a certain path in the meta-graph), and that child architecture is trained to optimize for $\taskObjective$. After the child architecture is trained, the corresponding nodes of the meta-graph are updated accordingly.


During the inference phase, $m(x; \theta_x) = C(x, G; \phi)$ is applied to each unseen input instance $x$. The generated $m(x; \theta_x)$ is an architecture that tailored for each $x$ and best trade-offs between $\taskObjective$ and $\archObjective$. Note that the latencies we reported in Section~\ref{section:experiments} has included the controller latency, since the controller is applied for each inference.



\subsection{Meta-Graph}

\label{section:method:meta-graph}

Meta-graph is a weight-sharing mechanism designed to represent the search space of all possible child architectures with two important properties: (a) any end-to-end path (from source to sink) within the meta-graph is a valid child architecture, and (b) the performance (\eg, accuracy or latency) of this child architecture, without any further training, serves as a good proxy for the final performance (\ie, fully-trained performance). Without using the meta-graph, a straightforward approach of constructing instance-aware classifier might be: train many models, then introduce a controller to assign each input instance to the most suitable model. This approach is not feasible since the total number of parameters in the search space grows linearly w.r.t. the number of models considered, which is usually a very large number; for example, in this work, the search space contains $10^{25}$ child architectures. Therefore, \modelName adapts the meta-graph to reduce the total number of parameters via weight sharing; specifically, if two child architectures share any part of the meta-graph, only one set of parameters required to represent the shared sub-graph.

Next, we explain how the meta-graph is pre-trained. At the beginning of every training iteration, part of the meta-graph is randomly zero out (also called ``drop-path'' in \cite{bender2018understanding}), and the rest of modules within the meta-graph forms a child architecture. Then this child architecture is trained to optimize $\taskObjective$ (\eg, classification accuracy) and updates the weights of the corresponding part of the meta-graph. Note that the ``drop-path'' rate is a hyper-parameter between [0, 1]. The drop-path rate that the meta-graph trained with will affect how the controller explores the search space in the early stage. In this work, we achieve good results by linearly increasing the drop-path rate from the middle of pre-training and eventually reach to 50\%.

\subsection{Controller}
\label{section:method:agent}

\modelName controller is different to the one in conventional NAS that aims at training for effectively exploring in the search space. Given an input image, the \modelName controller proposes a child architecture by $m(x; \theta_x) = C(x; \phi)$. Therefore, during the inference phase, the controller is still required, and the design principle of the controller is to be fast since its latency is included as part of the inference procedure. The controller is responsible for capturing the low-level representations (\eg, the overall color, texture complexity, and sample difficulty) of each instance, then dispatches the instance to the proposed child architecture that can be treated as the domain expert to make accurate decisions. In this work, we use a three-layer convolutional network with large kernels as the implementation of a \modelName controller. Qualitative analysis and visualizations of how the controller categorizes samples are provided in Section~\ref{section:experiments:qualitative} (see Figure~\ref{fig:agent-categorize-c10} for example).

Next, we elaborate on the exploration strategy and reward function to train the controller. We also introduce a technique ``policy shuffling'' to stabilize the joint training.

\paragraph{Exploration Strategy.}
We formulate each architecture to be a set of binary options indicating whether each convolutional kernel within the meta-graph is selected. The controller takes each input image and generates a probability vector $\boldsymbol{p}$ indicating the probability of selecting a certain convolutional kernel. Then Bernoulli sampling is applied to this probability vector for exploring the architecture space. We adopt the entropy minimization mentioned in~\cite{mnih2016asynchronous}, which improves exploration by encouraging a more deterministic policy (either select or not select a kernel). To further increase the diversity of sampling result during exploring the architecture space, we adopt the encouragement parameter $\alpha$ described in \cite{wu2018blockdrop} which mutates the probability vector by $\boldsymbol{p}' = \alpha \cdot \boldsymbol{p} + ( 1 - \alpha) \cdot ( 1 - \boldsymbol{p} )$. Note that to ensure the shape of feature maps to be correct, we enforce at least one module to be selected at each layer.

The controller is trained with policy gradient. Similar to training procedure proposed in ~\cite{zoph2016neural, wu2018blockdrop}, we introduce a ``self-critical'' baseline~\cite{rennie2017self} $R(\widetilde{p})$ to reduce the variance of instance-wise reward $R(p')$, where $\widetilde{p}_{i} = 1$ if $p > 0.5$, and $\widetilde{p}_{i} = 0$ otherwise. The policy gradient is estimated by:
\begin{equation}
    \nabla_{\phi} J = \expectation [ (R(p')-R(\widetilde{p})) \nabla_{\phi} \sum_i \text{log} \, P(a_i) ] \, ,
\end{equation} 
which $\phi$ is the parameters of the controller and each $a_i$ is sampled independently by a Bernoulli distribution with respect to $p_i \in \boldsymbol{p}$.

\paragraph{Reward Function.}
The reward function is designed to be multi-objective that takes both $\taskObjective$ and $\archObjective$ into account. The reward is calculated as:
\begin{equation}
    R = 
    \begin{cases}
         R_T \cdot R_A & \text{if $R_T$ is positive,} \\
         R_T & \text{otherwise},
    \end{cases}
\end{equation} 
which $R_T$ and $R_A$ are obtained from $\taskObjective$ and $\archObjective$. The design of $R$ is based on the observation that $\taskObjective$ is generally more important and preferred than $\archObjective$. As a result, $\archObjective$ is only taken into account when $\taskObjective$ is secured. Otherwise, the controller is first ensured to maximize $R_T$. Even for the cases where $R_T$ is positive, $R_A$ is treated to be ``preferred'' (not enforced), which is done by normalizing $R_A$ to the range $[0, 1]$ that becomes a discounting factor to $R_T$ and never provides negative penalties to the controller through policy gradient.

Another observation is that optimizing $\archObjective$ is generally challenging to optimize, which at times collapses the controller training. One possible reason is: take $R_T$ to be accuracy and $R_A$ to be latency as an example, architectures with both good latency and desirable accuracy are rare. Meanwhile, our ``instance aware'' setting collects reward in a ``instance-wise'' manner, finding architectures with extremely low latency for \textit{\textbf{all}} samples (trivially selecting most simple kernels) is significantly easier than having generally high accuracy for any sample. Therefore, in the early stage of the controller exploration, such pattern encourages the controller to generate shallow architectures and directly ignores accuracy. Eventually, the policy collapses to a single architecture with extremely low latency with a poor accuracy close to random guessing.



To address the aforementioned problem, we propose a training framework using ``dynamic reward.'' Dynamic reward encourages the controller to satisfy a gradually decreasing latency reward with bounds (upper-bound $U_t$ and lower-bound $L_t$, which $t$ is the number of epochs) during search time. The idea of dynamic reward shares a similar concept with curriculum learning~\cite{bengio2009curriculum}, except that we aim at gradually increasing the task difficulty to avoid the sudden collapsing. This is also similar in the spirit of \cite{wong2018transfer}, which accelerates NAS by transferring the previously learned policy to new tasks (the new reward bound in the case of \modelName). In this work, we propose the reward $R_A$ to be a quadratic function parametrized by $U_t$ and $L_t$. For each sample, we measure architecture-related performance $z$, then calculate $R_A = - \frac{1}{\gamma} \, (z-U_t) \times (z-L_t)$, which $\gamma$ is a normalization factor that normalizes $R_A$ to the range $[0, 1]$ by $\gamma = ( {\frac{U_t - L_t}{2}} )^2$. Such a design (quadratic function) encourages the controller to maintain the expectation of $\archObjective$ near the center of the reward interval, while still be aware of $\taskObjective$. Otherwise, the child architectures may fall outside the reward interval upon the reward interval changes.

\paragraph{Policy Random Shuffling.}
An ideal exploration behavior for the controller is to sample diverse and high-reward child architectures. At the same time, if the meta-graph is overly fine-tuned and trained by a subset of configurations for each sample(s), the exploration diversity will drop since the controller tends to focus on exploring a subset of child architectures that have been fine-tuned to have higher accuracy as the reward. To mitigate such an effect, we introduce random shuffling on policies: after the controller determines the policies (\ie, child architectures),  these policies are shuffled (by \textit{shuffling} ($\boldsymbol{p}'$) in each batch of training samples) and each policy will be re-paired to a sample within the same batch during meta-graph fine-tuning phase. This technique keeps a particular child architecture from memorizing a subset of samples and keeps the path preference into the meta-graph. Empirically, we observe that without random shuffling, the search process usually fails at the very beginning.



\section{Experiments}
\label{section:experiments}
In this section, we explain and analyze the building blocks of \modelNamePunc. We start by demonstrating some quantitative results of \modelName against other models. Then we visualize and discuss some empirical insights of \modelNamePunc. Throughout the experiments, we use the same search space described in Section~\ref{section:experiments:setup}. We specify our main task to be image classification, though \modelName is expected to work for most vision tasks if given sufficient computation power. For the search objectives, we choose accuracy as our task-dependent objective and latency as the architecture-dependent objective, which are the most influential factors of architecture choice in real-world applications.

\subsection{Experiment Setups}
\label{section:experiments:setup}


\begin{figure}[t]
\centering
\includegraphics[width=1\linewidth]{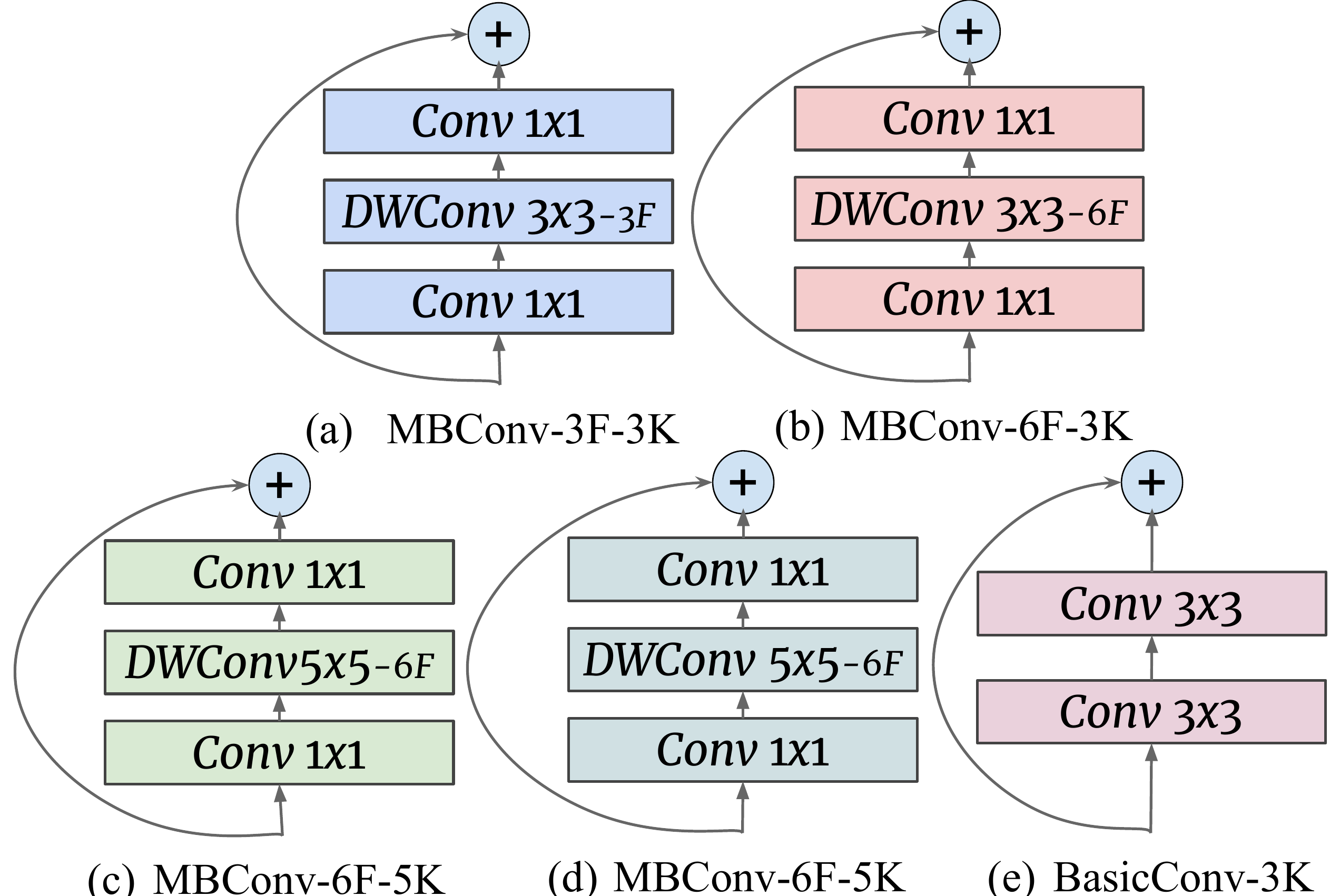}
\caption{The five module options in each cell of InstaNAS including basic convolution~\cite{he2016deep} and mobile inverted bottleneck convolution~\cite{sandler2018mobilenetv2} with different expansion ratios (F) and kernel sizes (K). Note that DWConv stands for depth-wise convolution.} 
\label{fig.searchspace}

\end{figure}

\paragraph{Search Space.}
We validate \modelName in a search space inspired by~\cite{tan2018mnasnet}, using MobileNetV2 as the backbone network. Our search space consists of a stack of 17 cells and each cell has five module choices as shown in Figure~\ref{fig.searchspace}. Specifically, we allow one basic convolution \textit{(BasicConv)} and four mobile inverted bottleneck convolution \textit{(MBConv)} layers with various kernel sizes \{3, 5\} and filter expansion ratios \{3, 6\} as choices in the cell, which equals to $2^{5}=32$ possible combinations. Different from~\cite{dong2018dpp, zoph2017learning}, we do not restrict all cells to share the same combination of architectures. Therefore, across the entire search space, there are approximately $32^{17}\simeq10^{25}$ child architecture configurations. Note that for cells that consist of stride two convolutions or different numbers of input/output filter size, at least one module has to be selected (Module \textit{(a)} in default).

\begin{figure}[t]
\centering
\includegraphics[width=1\linewidth]{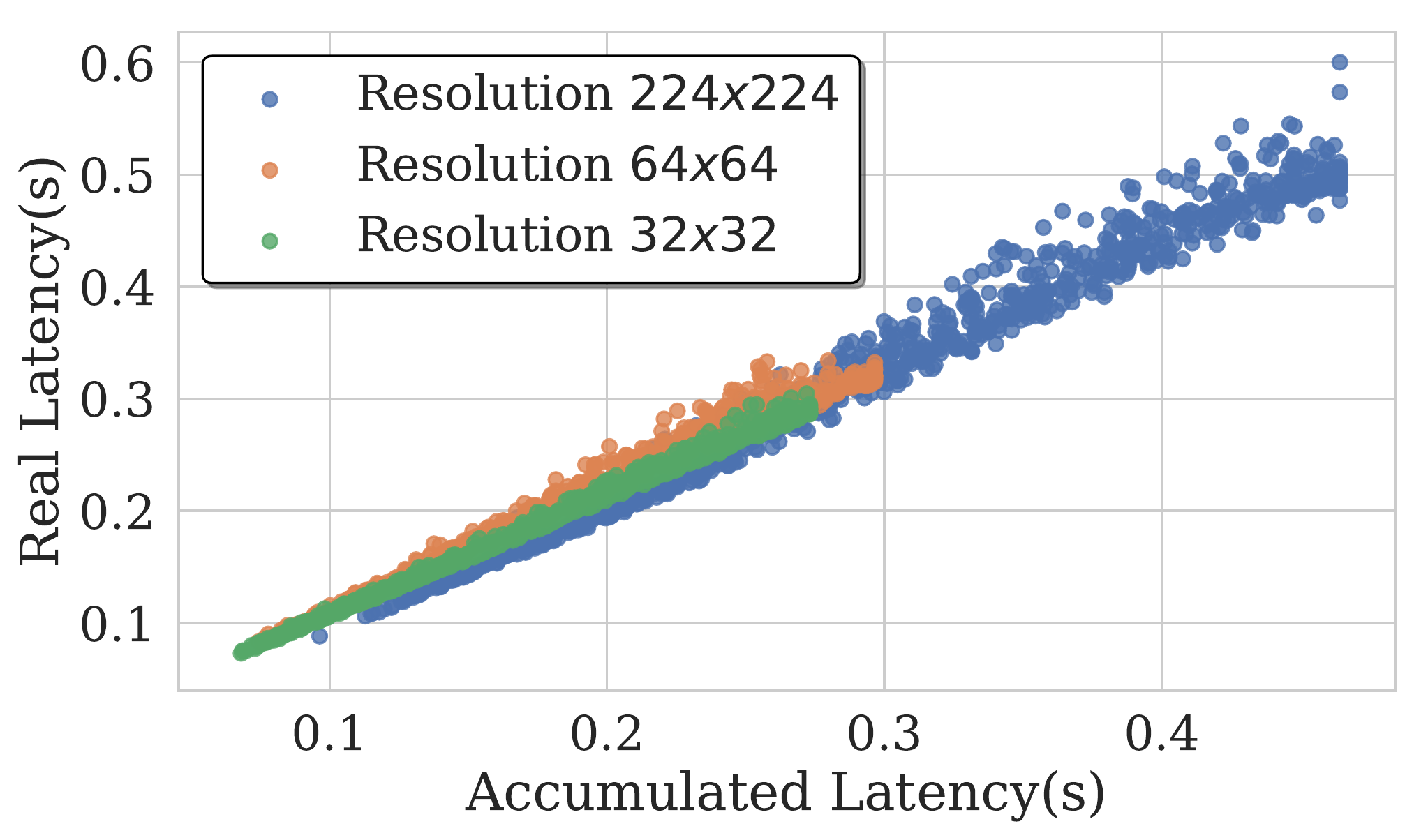}
\caption{The estimated latency (profile module-wise latency and accumulate the values) has a strongly positive correlation to the real latencies. In all three resolution settings, we sample 10,000 child architectures from the search space and measure the real latency and estimated latency.}
\label{fig.latency}
\end{figure}


\paragraph{Module Latency Profiling.}
In the instance-aware setting, evaluating the latency reward is a challenging task as each input instance is possibly assigned to different child architectures. However, measuring the latency individually for each child architecture is considerably time costly during training. Therefore, to accelerate training, we evaluate the latency reward with estimated values. Specifically, we build up module-wise look-up tables with pre-measured latency consumption of each module. For each sampled child architecture, we look up the table of each module and accumulate the layer-wise measurements to estimate the network-wise latency consumption. Figure~\ref{fig.latency} compares the estimated latency (the sum of module-wise latency from the module-wise look-up tables) and the real latency on a workstation with Intel Core i5-7600 CPU. The result shows real and estimated latency numbers are highly correlated: the overall correlation coefficient is 0.97, 0.98, and 0.97 for input resolution 32, 64, and 224, respectively. This error rate is considerably small, which shows that our estimated value is a good approximation for real latency. 


\subsection{Quantitative Results}
\label{section:experiments:quantitative}

\begin{table}[]
\small
\centering
\begin{tabular}{lccr}
\toprule
Model & Err. (\%) & Latency \\ \midrule
ResNet50\cite{he2016deep} & 6.38 & 0.051 $\pm$ 0.003 \\
ResNet101\cite{he2016deep}& 6.25 & 0.095 $\pm$ 0.002 \\
ShuffleNet v$2$ $1.0\times$ \cite{ma2018shufflenet} & 7.40 & 0.035 $\pm$ 0.001 \\
ShuffleNet v$2$ $1.5\times$ \cite{ma2018shufflenet} & 6.36 & 0.052 $\pm$ 0.002 \\
IGCV$3$-D $1.0\times$ \cite{sun2018igcv3} & 5.54 & 0.185 $\pm$ 0.003 \\
IGCV$3$-D $0.5\times$ \cite{sun2018igcv3} & 5.27 & 0.095 $\pm$ 0.006 \\ \midrule
NASNet-$A$ \cite{zoph2017learning} & 3.41 & 0.219 $\pm$ 0.006 \\
DARTS \cite{liu2018darts} & 2.83 & 0.236 $\pm$ 0.004 \\
DPP-Net-$Mobile$ \cite{dong2018dpp} & 5.84 & 0.062 $\pm$ 0.004  \\ \midrule
MobileNet v$2$ $0.4\times$ \cite{sandler2018mobilenetv2} & $7.44$ & $0.038$ $\pm$ 0.003 \\
MobileNet v$2$ $1.0\times$ \cite{sandler2018mobilenetv2} & $5.56$ & $0.092$ $\pm$ 0.002 \\
MobileNet v$2$ $1.4\times$ \cite{sandler2018mobilenetv2} & $4.92$ & $0.129$ $\pm$ 0.002 \\ \midrule
InstaNAS-C10-$A$ & \textbf{4.30} & 0.085 $\pm$ 0.006 \\
InstaNAS-C10-$B$ & \textbf{4.50} & 0.055 $\pm$ 0.002 \\
InstaNAS-C10-$C$ & 5.20 & \textbf{0.047} $\pm$ 0.002 \\
InstaNAS-C10-$D$ & 6.00 & \textbf{0.033} $\pm$ 0.001 \\
InstaNAS-C10-$E$ & 8.10 & \textbf{0.016} $\pm$ 0.006 \\
\bottomrule
\end{tabular} 
\caption{\modelName shows competitive latency and accuracy trade-off in CIFAR-10~\cite{krizhevsky2009learning} against other state-of-the-art human-designed models (first row) and NAS-found models (second row). All five \modelName models are all obtained within a single search, and the controller latency is already included in the reported latency. Note that we measure the model's error rates with our implementation if it is not reported in the original paper (\eg,~\cite{ma2018shufflenet,sandler2018mobilenetv2, tan2018mnasnet}). }
\label{table:performance_compare}
\end{table}

\paragraph{Experiments on CIFAR-10/100.}
We validate \modelName on CIFAR-10/100 with the search space described in the previous section. Across all training stages, we apply random copping, random horizontal flipping, and cut-out~\cite{devries2017improved} as data augmentation methods. For pre-training the meta-graph, we use Stochastic Gradient Descent optimizer with initial learning rate 0.1. Each training batch consists of 32 images on a single GPU. After the joint training ends, some controllers are picked by human preference by considering the accuracy and latency trade-off. At this point, the accuracy measured in the joint training stage can only consider as a reference value, the meta-graph needs to re-train from scratch with respect to the picked policy. We use Adam optimizer with learning rate 0.01 and decays with cosine annealing. More detail is provided in supplementary.



\begin{table}[t]
\small
\centering
\begin{tabular}{lccr}
\toprule
Model & Err. (\%) & Latency  \\ \midrule
ShuffleNet v$2$ $0.5\times$ \cite{ma2018shufflenet} & 34.64 & 0.016 $\pm$ 0.001  \\
ShuffleNet v$2$ $1.0\times$ \cite{ma2018shufflenet} & 30.60 & 0.035 $\pm$ 0.001  \\
ShuffleNet v$2$ $1.5\times$ \cite{ma2018shufflenet} & 28.30 & 0.052 $\pm$ 0.004  \\\hline
MobileNet v$2$ $0.4\times$ \cite{sandler2018mobilenetv2} & 30.72 & 0.049 $\pm$ 0.068  \\
MobileNet v$2$ $1.0\times$ \cite{sandler2018mobilenetv2} & 27.00 & 0.092 $\pm$ 0.002  \\
MobileNet v$2$ $1.4\times$ \cite{sandler2018mobilenetv2} & 25.66 & 0.129 $\pm$ 0.007  \\ \hline
InstaNAS-C100-$A$ & \textbf{24.20} & 0.089 $\pm$ 0.003  \\
InstaNAS-C100-$B$ & \textbf{24.40} & 0.086 $\pm$ 0.006  \\
InstaNAS-C100-$C$ & \textbf{24.90} & 0.065 $\pm$ 0.003  \\
InstaNAS-C100-$D$ & 26.60 & \textbf{0.055} $\pm$ 0.004  \\
InstaNAS-C100-$E$ & 27.80 & \textbf{0.046} $\pm$ 0.004  \\
\bottomrule 
\end{tabular} 
\caption{InstaNAS consistently provides significant accuracy improvement and latency reduction on CIFAR-100. Again, all the InstaNAS variants are obtained within a single search.}
\label{table:performance_compare_cifar100}

\end{table}


Table~\ref{table:performance_compare} shows the quantitative comparison with state-of-the-art efficient classification models and NAS-found architectures. The result suggests \modelName is prone to find good trade-off frontier relative to both human-designed and NAS-found architectures. In comparison to MobileNetV2 ($1.0\times$), which the search space is referenced to, InstaNAS-C10-$A$ improves accuracy by 1.26\% without latency trade-off; InstaNAS-C10-$C$ reduces 48.9\% average latency with comparable accuracy, and InstaNAS-C10-$E$ reduces 82.6\% latency with moderate accuracy drop. Note that these three variances of \modelName are all obtained within a single search, then re-train from scratch individually.

Our results on CIFAR-100 are shown in Table~\ref{table:performance_compare_cifar100}, which the average latency consistently shows reduction - with 40.2\% comparing to MobileNetV2 $1.0\times$, 36.1\% comparing to ShuffleNetV2 $2.0\times$. InstaNAS stably shows overall improvement in the trade-off frontier against competitive state-of-the-art models. 

\begin{table}[h]
    \small
    \centering
    \begin{tabular}{lccr}
        \toprule
        Model & Err. (\%) & Latency \\
        \midrule
        MobileNetV1 & 56.4 & - \\
        MobileNetV2 1.0 & 48.1 & 0.264 $\pm$ 0.012 \\
        MobileNetV2 1.4 & 42.8 & 0.377 $\pm$ 0.006  \\\hline
        InstaNAS-Tiny-A & \textbf{\underline{41.4}} & 0.223 $\pm$ 0.005  \\
        InstaNAS-Tiny-B & 43.9 & 0.179 $\pm$ 0.007 \\
        InstaNAS-Tiny-C & 46.1 & \textbf{\underline{0.171}} $\pm$ 0.007 \\
        \bottomrule
    \end{tabular} 
    \caption{InstaNAS can generalize to larger scale dataset and provide decent latency on TinyImageNet. MobileNetV1 result on TinyImageNet is also included as a reference~\cite{kim2018munet}.}


\label{table:tiny-imagenet}
\end{table}

\begin{figure}[t]
\centering
\includegraphics[width=1\linewidth]{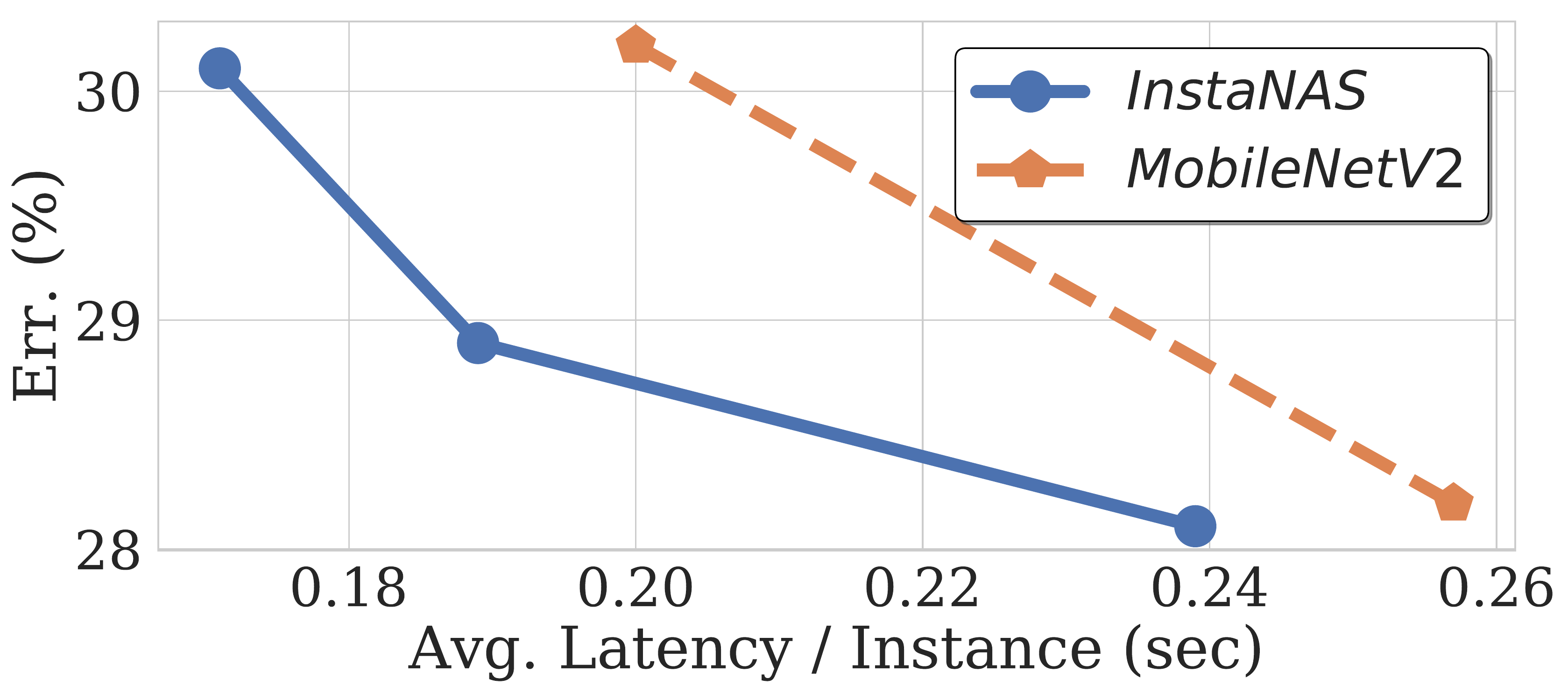}%

\caption{ImageNet results show that InstaNAS can consistently provides latency reduction with competitive accuracy against MobileNetV2.}
\label{fig.imgnet}

\end{figure}

\paragraph{Experiments on TinyImageNet and ImageNet.}
To validate the scalability, stability and generalization of InstaNAS, we evaluate our approach on two more fine-grained datasets, TinyImageNet and ImageNet. We ran the experiment using directly the same set of hyperparameters configuration from the CIFAR-10/100 experiment. As shown in Table~\ref{table:tiny-imagenet} and Figure~\ref{fig.imgnet}, InstaNAS comparing to MobileNetV2, again, found accurate model with 35.2\% latency reduction  on TinyImageNet and 14.5\% on ImageNet. Furthermore, if moderate accuracy drop ($\simeq$ 0.7\%) is tolerable, InstaNAS can further achieve 26.5\% average latency reduction on ImageNet. We report InstaNAS search time and numbers of parameters in the supplementary material.

\begin{figure}[t]
\centering
\includegraphics[width=1\linewidth]{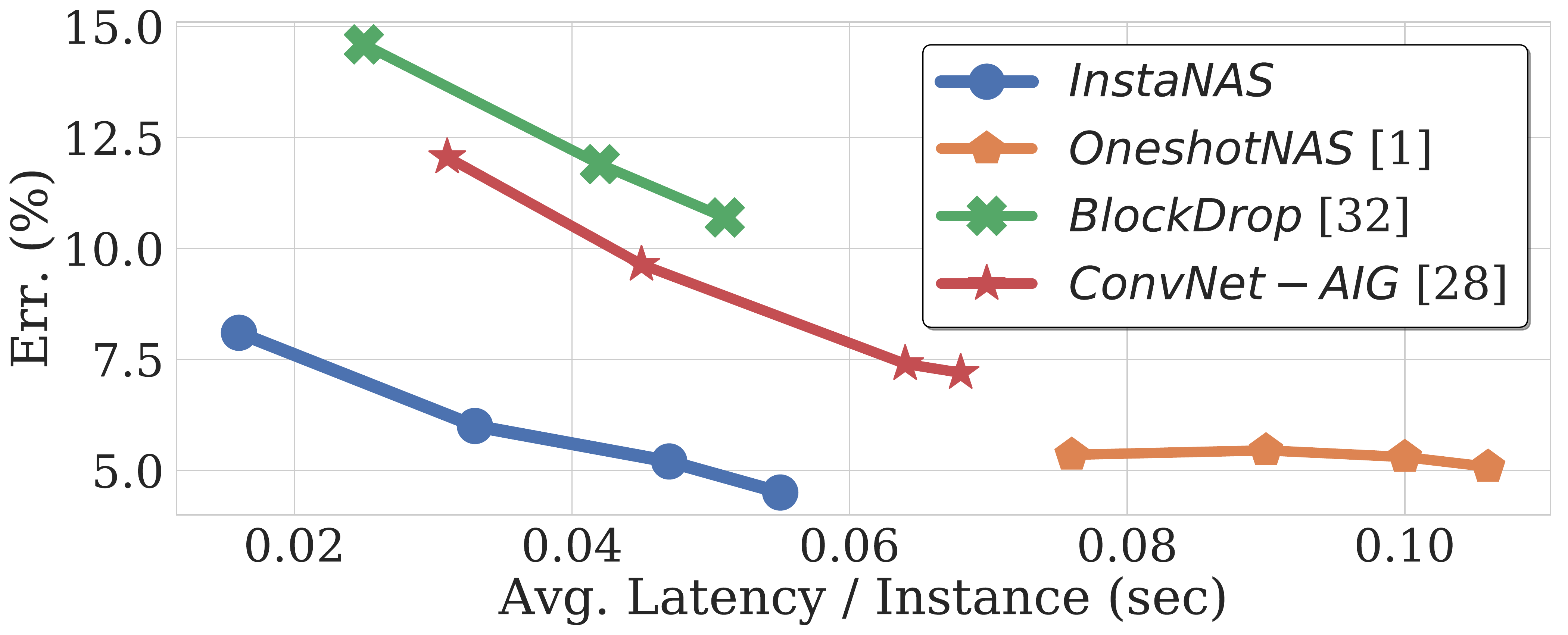}%

\caption{InstaNAS out-performs all the related baseline methods (\ie, one-shot architecture search~\cite{bender2018understanding} and other state-of-the-arts conditional computing methods~\cite{veit2018convolutional, wu2018blockdrop}) within MobileNetV2 search space.}
\label{fig.compare}

\end{figure}

\paragraph{Comparison with state-of-the-arts.}
In this section, we compare and show that InstaNAS outperforms several state-of-the-art search methods. Figure~\ref{fig.compare} illustrates the best architectures found on the trade-off (accuracy v.s. latency) frontier for \modelName and several state-of-the-arts: OneshotNAS~\cite{bender2018understanding}, BlockDrop~\cite{wu2018blockdrop} and ConvNetAIG~\cite{veit2018convolutional}. We follow the one-shot search procedures~\cite{bender2018understanding} to sample 10,000 models from the meta-graph and train the trade-off frontier points from scratch on CIFAR-10. From Figure~\ref{fig.compare}, we observe that the trade-off frontier achieved by \modelName is significantly better than OneshotNAS and other methods; note that the ideal curve should be closer to bottom-left corner, meaning the architectures found are accurate and fast.


Furthermore, compared to BlockDrop~\cite{wu2018blockdrop} and ConvNet-AIG~\cite{veit2018convolutional}, the architectures found by \modelName have both higher accuracy and lower latency, which dominates the conditional computing models and in turn confirms instance-awareness to be an effective characteristic for NAS.

\begin{figure*}[!t]
\centering
\includegraphics[width=0.96\linewidth]{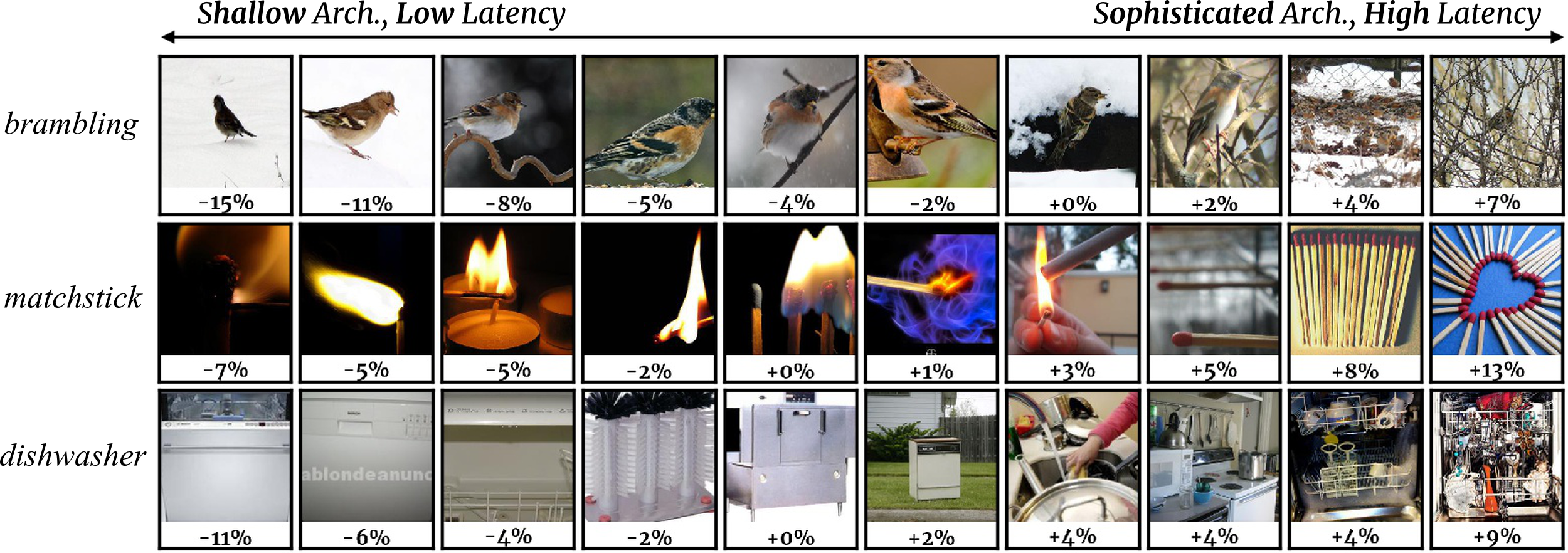}%
\caption{\modelName selects architectures tailored for each image. Each row represents samples from ImageNet with the same label; the images on the left are considered to be ``simpler'' and images on the right are ``complex.'' These levels are determined by the controller, which also matches humans' perception: \eg, cluttered background, high intra-class variation, illumination conditions. The number below each image represents the relative difference on latency. 0\% means the average latency of all architectures selected for the images within certain class. See supplementary materials for more samples.}
\vspace{-1em}
\label{fig:slowfast}
\vspace{-0.5em}
\end{figure*}

\begin{figure}[h]

  \includegraphics[width=\linewidth]{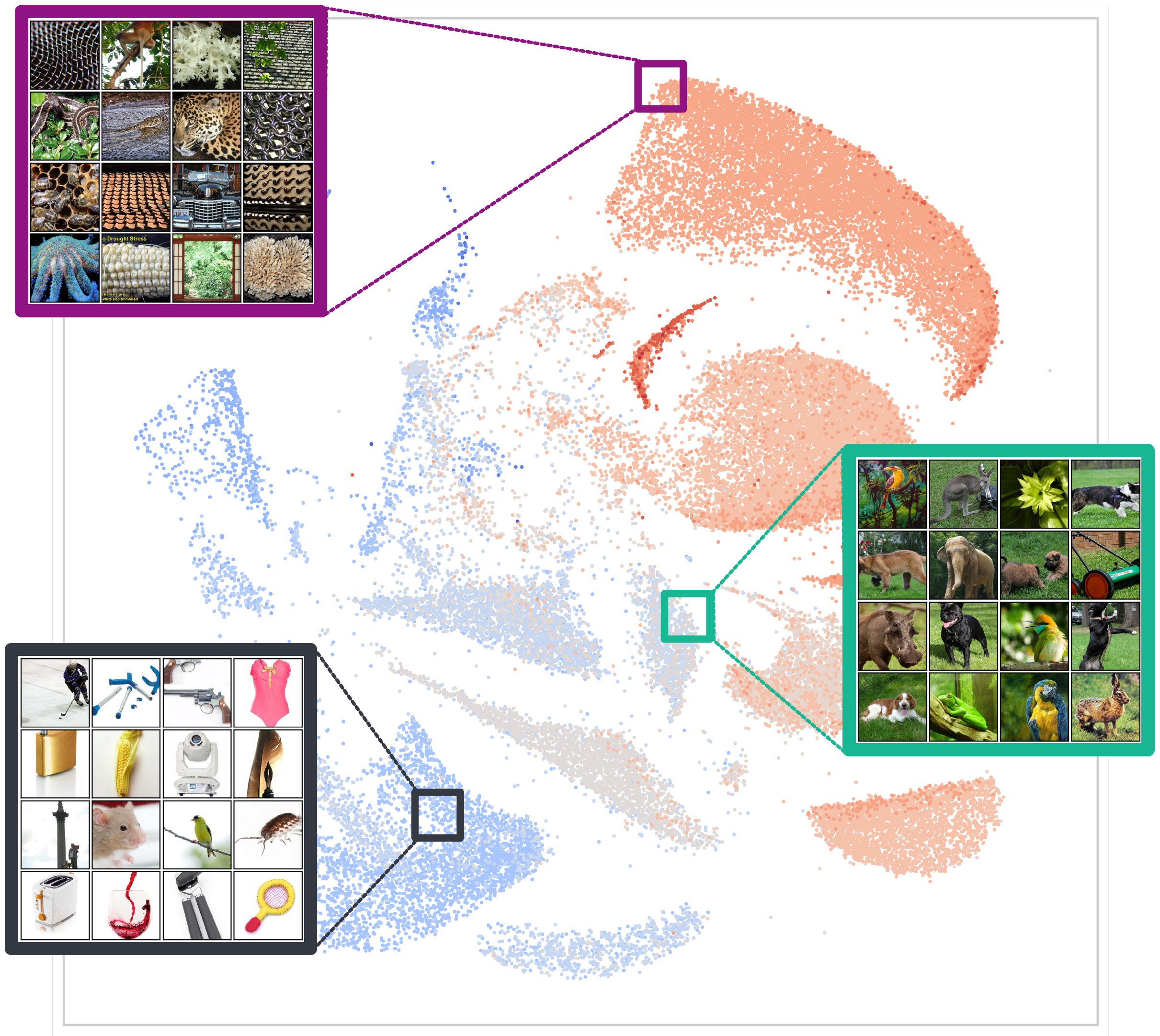} 

  \caption{Distribution of InstaNAS architectures on ImageNet. Each point corresponds to an architecture probability vector $p$. We adopt UMAP~\cite{umap,umap-software} to project high-dimensional $p$ into 2D space, and color-code each architecture by its inference latency: \textit{\textcolor{red}{red}} for high latency and \textit{\textcolor{blue}{blue}} for low latency. We also visualize three set of instances (in rectangle boxes) and instances in each box share the same architecture. Notice that the controller categorizes input instances base on their low-level visual characteristic, such as the background color (\textit{\textcolor{green}{green}}), object position/orientation (\textit{\textbf{black}}) and texture complexity (\textit{\textcolor{purple}{purple}}). Then the controller assigns each instance to a down-stream expert architecture for accurate classification.}
  \label{fig:agent-categorize-c10}
  \vspace{-0.5em}
\end{figure}

\subsection{Qualitative Results}
\label{section:experiments:qualitative}
In this section, we provide a qualitative analysis on the child architectures selected by \modelName for ImageNet. Figure~\ref{fig:slowfast} illustrates the various images of three classes (brambling, matchstick, and dishwasher) sorted by its assigned architecture's latency (showed as the number below each image) normalized by the average latency---0\% represents the average latency of all the architectures assigned to the images under a certain class. The images on the left are considered to be ``simpler'' (the architectures used have lower latency), and images on the right are ``complex.'' Note that these levels are determined by the controller, which also matches humans' perception on the image complexity: \eg, images with a cluttered background, high intra-class variation, illumination conditions are more complex and therefore sophisticated architectures (with higher latency) are assigned to classify these complex images. 

\footnotetext{More examples with complete CIFAR-10 and ImageNet validation results are provided in \url{https://goo.gl/Gx6nos}. Each folder corresponds to a unique child architecture selected by the controller.}

Figure~\ref{fig:agent-categorize-c10} illustrates architecture distribution (projected onto 2-D) with each dot representing an architecture and the color being the corresponding latency (red represents high latency and blue means low). We also randomly select three architectures and highlight (in three color-coded boxes) the images samples assigned to them (by the controller) for making an inference. Notice that the image samples in each box share similar low-level visual patterns (\eg, background color, object position/orientation, and texture complexity) that agree with humans' perception. Both qualitative analyses confirm \modelName's design intuition that the controller learns to discriminate each instance based on its low-level characteristic (that agrees with humans' perception) for best assigning that instance to the corresponding expert architecture.




\section{Conclusion and Future Works}
\label{section:conclusion}
In this paper, we propose InstaNAS, the first instance-aware neural architecture search framework. InstaNAS exploits instance-level variation to search for a \textit{distribution of architectures}; during the inference phase, for each new image InstaNAS selects one corresponding architecture that best classifies the image while using less computational resource (\eg, latency). The experimental results on CIFAR-10/100, TinyImageNet, and ImageNet all confirm that better accuracy/latency trade-off is achieved comparing to MobileNetV2, which we designed our search space against. Qualitative results further show that the proposed instance-aware controller learns to capture the low-level characteristic (\eg, difficulty, texture and content style) of the input image, which agrees with human perception.



One important future work direction is to reduce the total number of parameters of InstaNAS. Although only a portion of modules is activated for each inference, considering system stableness, the full meta-graph still needs to be loaded into the memory. Despite this may not be an issue for some latency-oriented scenarios with an adequate amount of memory (\eg, self-driving car and cloud service), this problem still becomes a drawback for deploying \modelName on edge devices or chips. How to further reduce the total parameters of \modelName is an important next step. Another potential solution for such a problem may be restricting the controller from switching policies for some special applications that have a correlation between consecutive inferences (\eg, surveillance camera, on-board camera, and drone).


\vspace{1em}


\section*{Acknowledgements}

We are grateful to the National Center for High-performance Computing for computer
time and facilities, and Google Research, MediaTek, MOST 107-2634-F007-007
for their support. This research is also supported in part by the Ministry
of Science and Technology of Taiwan (MOST 107-2633-E-002-001), MOST Joint Research Center for AI Technology and All Vista Healthcare, National Taiwan
University, Intel Corporation, and Delta Electronics. We would like to thank Jin-Dong Dong and Wei-Chung Liao for their helpful comments
and discussions.

\end{document}